\documentclass{article}

\usepackage{arxiv}

\usepackage[utf8]{inputenc} 
\usepackage[T1]{fontenc}    
\usepackage{hyperref}       
\usepackage{url}            
\usepackage{booktabs}       
\usepackage{amsfonts}       
\usepackage{nicefrac}       
\usepackage{microtype}      
\usepackage{lipsum}		
\usepackage{graphicx}
\usepackage{natbib}
\usepackage{doi}

\usepackage{xcolor}         

\usepackage{multirow}
\usepackage{multicol}
\usepackage{makecell}
\usepackage[linesnumbered, ruled]{algorithm2e}
\usepackage{wrapfig}
\usepackage{amssymb}
\usepackage{amsmath}
\usepackage{lineno}
\usepackage{setspace}
\usepackage{caption}
\let\oldgather\gather
\let\oldendgather\endgather
\renewenvironment{gather}{\linenomathNonumbers\oldgather}{\oldendgather\endlinenomath}

\title{Vertical Federated Learning Hybrid Local Pre-training}

\author{ {\hspace{1mm}Wenguo Li} \\
	Zhejiang Lab\\
	Kechuang Road 1, Hangzhou, 311121, China\\
	\texttt{liwg@zhejianglab.com} \\
	\And
	{\hspace{1mm}Xinling Guo} \\
	Zhejiang Lab\\
	Kechuang Road 1, Hangzhou, 311121, China\\
	\texttt{gxl@zhejianglab.com} \\
    \And
	{\hspace{1mm}Xu Jiao} \\
	Zhejiang Lab\\
	Kechuang Road 1, Hangzhou, 311121, China\\
	\texttt{jiaoxu@zhejianglab.com} \\
    \And
    {\hspace{1mm}Tiancheng Huang} \\
	School of Computer Science and Engineering\\
	Nanyang Technological University\\
	\texttt{tiancheng.huang@ntu.edu.sg} \\
     \And
	{\hspace{1mm}Xiaoran Yan} \\
	Zhejiang Lab\\
	Kechuang Road 1, Hangzhou, 311121, China\\
	\texttt{yanxr@zhejianglab.com} \\
     \And
	{\hspace{1mm}Yao Yang} \\
	Zhejiang Lab\\
	Kechuang Road 1, Hangzhou, 311121, China\\
	\texttt{yangyao@zhejianglab.com} \\
}

\renewcommand{\shorttitle}{\textit{arXiv} Template}

\begin{document}
\maketitle

\begin{abstract}
Vertical Federated Learning (VFL), which has a broad range of real-world applications, has received much attention in both academia and industry. Enterprises aspire to exploit more valuable features of the same users from diverse departments to boost their model prediction skills. VFL addresses this demand and concurrently secures individual parties from exposing their raw data. However, conventional VFL encounters a bottleneck as it only leverages aligned samples, whose size shrinks with more parties involved, resulting in data scarcity and the waste of unaligned data. To address this problem, we propose a novel VFL Hybrid Local Pre-training (VFLHLP) approach. VFLHLP first pre-trains local networks on the local data of participating parties. 
Then it utilizes these pre-trained networks to adjust the sub-model for the labeled party or enhance representation learning for other parties during downstream federated learning on aligned data, boosting the performance of federated models.
The experimental results on real-world advertising datasets, demonstrate that our approach achieves the best performance over baseline methods by large margins. The ablation study further illustrates the contribution of each technique in VFLHLP to its overall performance.
\end{abstract}


\section{Introduction}
Federated learning \cite{li2020federated,zhang2021survey} is a paradigm of distributed machine learning that allows multiple parties to collaboratively train a joint model while preserving their own data privacy. Vertical federated learning (VFL) \cite{liu2022vertical, wei2022vertical} is one of the typical federated learning settings, specifically for the scenario where participants share the same or partially aligned samples but different features. This scenario is widely encountered in practice, such as in finance, health care, telecommunications, and other domains \cite{zheng2020vertical, liu2019confederated, zhang2020vertical}. VFL provides a legal and applicable solution for multiple enterprises to improve their model performance by leveraging valuable feature or label information from other collaborators.
Therefore, VFL research has been drawing increasing attention in both academia and industry in recent years.

\begin{figure}[h]
    \centering
    \includegraphics[width=0.9\linewidth]{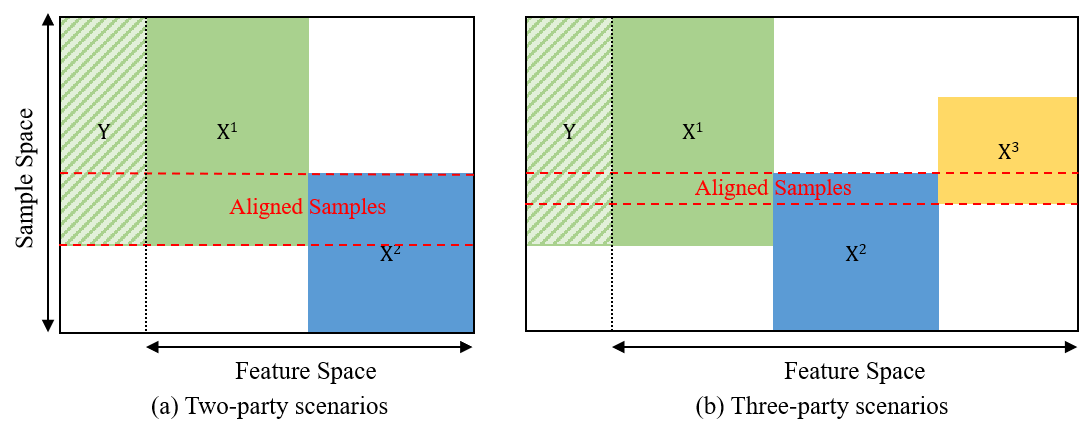}
    \caption{The illustration of data partitioning in VFL scenarios. Each party holds different features, and Party 1 owns a fixed amount of labels. The aligned samples are bounded with red dashed lines, whose number occupies only a small portion of individual party samples and gets shrunk when more parties join in training.}
    \label{fig:1}
\end{figure}

However, conventional VFL is only able to leverage aligned samples. In reality, aligned samples are always limited, and their amount decreases sharply as the number of participating parties increases (see Figure \ref{fig:1}), leading to a dramatic decline in the performance of federated models, which cannot satisfy industrial prediction demand. We refer to this as the few-overlap problem (specifically on the sample space) in real-world VFL applications.
To address this problem, \cite{kang2022fedcvt} proposed a semi-supervised learning approach, Federated Cross-View Training (FedCVT), to estimate the information of missing features and labels for downstream supervised VFL on the full sample-feature space. 
This approach essentially depends on the accurate prediction of missing data and is limited to the scenarios of two parties. 
\cite{castiglia2022self, he2022hybrid} integrated self-supervised learning (SSL) into VFL to enhance the capability of representation learning via exploiting local features of individual parties. 
Nevertheless, these works only utilize the labels of aligned samples, causing label underutilization of unaligned ones. In realistic scenarios, the institution with labels has a fixed number of labeled samples regardless of how many parties participate in federated training, and the amount of labels is often much larger than the aligned sample size, especially as the number of parties grows. 
Existing VFL approaches rarely fully exploit available local information in multi-party scenarios. 

To fill these gaps, we propose a VFL Hybrid Local Pre-training (VFLHLP) approach. VFLHLP exploits local knowledge derived from all accessible local data of participating parties to improve joint learning on limited aligned samples. 
We underline the distinction in label ownership among participating parties and thus adopt different strategies. 
Specifically, we leverage the idea of knowledge transfer for the label owner to exploit both labels and features of unaligned samples while applying a specific SSL technique for multiple collaborators to address the feature waste of unaligned ones.
Additionally, we concentrate our study on tabular datasets as it is the most common data type in real-world VFL scenarios.
Our key contributions are as follows:

\begin{itemize}
    \item We propose a method named VFLHLP that makes full use of all available data by implementing knowledge transfer and SSL techniques to address the few-overlap issue widely encountered in real-world VFL applications.
    \item VFLHLP contains two stages of local pre-training and downstream VFL. As the pre-training stage is performed locally, the efficiency of federated training is scarcely compromised.
    \item We demonstrate our approach VFLHLP outperforms baselines by large margins in real-world advertising datasets. The ablation study illustrates the contribution of each technique to the overall performance boost.
\end{itemize}

\section{Related Work}

\subsection{Vertical Federated Learning}

Recent literature has explored various aspects of VFL concerning effectiveness, efficiency, and privacy. Many researchers have made efforts to improve efficiency in VFL by mitigating communication overhead via reducing coordination cost \cite{liu2019communication, zhang2021asysqn, khan2022communication} or compressing transmitted data \cite{li2020efficient, xu2021efficient, castiglia2022compressed} between parties. Other VFL studies concentrate on enhancing the effectiveness of local predictors \cite{li2022semi, sharma2019secure} or federated models \cite{feng2022vertical, he2022hybrid, yang2022multi} in limited aligned sample scenarios. Knowledge distillation \cite{li2022semi, wang2022vertical} and transfer learning \cite{sharma2019secure, feng2022semi} techniques have been applied to leverage aligned samples to improve local predictions. Self-supervised and semi-supervised learning approaches have been introduced to use local unaligned data to enhance conventional VFL models \cite{he2022hybrid, kang2022fedcvt}. Additional VFL research topics such as privacy, data valuation, and fairness are summarized in \cite{liu2022vertical}.

\subsection{Self (Semi)-Supervised Learning in VFL}
Self (Semi)-Supervised Learning \cite{jaiswal2020survey, liu2021self} makes use of abundant unlabeled data to alleviate poor model performance caused by label deficiency. Several studies have incorporated these techniques into VFL frameworks. FedCVT \cite{kang2022fedcvt}, leverages estimated representations for missing features and predicted pseudo-labels for unlabeled samples to expand the training set. It trains three classifiers jointly based upon different views of the expanded training set to improve VFL model performance. However, FedCVT is limited to two-party scenarios and may induce confirmation bias \cite{arazo2020pseudo}.
SS-VFL \cite{castiglia2022self} introduced self-supervised learning (SSL) into VFL, to improve the representation learning capability of federated networks on small aligned sample sets. 
\cite{he2022hybrid} proposed a Federated Hybrid Self-Supervised Learning framework (FedHSSL) that exploits cross-party views of aligned samples and local views of local samples and further aggregates invariant features shared among parties. However, these studies only consider that aligned samples own labels, unlike real cases where label owners have a fixed and much larger number of labeled samples than aligned ones.

\subsection{SSL for Tabular Datasets}
Self-supervised methods have achieved state-of-the-art results on images, text, and other homogeneous data. Adapting them to heterogeneous tabular data remains an open challenge. Recent studies have proposed various approaches to applying self-supervision to tabular data. VIME \cite{yoon2020vime} introduces auxiliary pretext tasks of reconstructing and estimating mask vectors from corrupted tabular data. The pretext tasks act as an additional form of supervision. TabNet \cite{arik2021tabnet} uses sequential attention mechanisms to determine the most salient features to focus on at each step. This avoids reasoning from all features at once. SubTab \cite{ucar2021subtab} formulates tabular learning as a multi-view representation problem. Input features are divided into distinct subsets to produce diverse views. Scarf \cite{bahri2021scarf} creates different partial views of the data by randomly corrupting subsets of features. A contrastive loss compares representations between these views. Nonetheless, extending these ideas to federated scenarios remains unexplored.

\section{Preliminaries}
Federated learning is comprised of three categories: Horizontal Federated Learning (HFL), Vertical Federated Learning (VFL), and Federated Transfer Learning (FTL) \cite{yang2019federated}. Distinguished from shared sample space in VFL, HFL refers to the scenario where multiple parties share the same feature space but different sample space, and FTL refers to the scenario where both feature and sample spaces of individual parties have limited overlaps. More clearly, they are summarized in former literature \cite{yang2019federated} as follows:

\textbf{Horizontal Federated Learning (HFL):} 
\begin{equation}
    \mathcal X_i=\mathcal X_j, \mathcal Y_i=\mathcal Y_j, \mathcal I_i\neq\mathcal I_j, \forall \mathcal D_i,\mathcal D_j,i \neq j
\end{equation}

\textbf{Vertical Federated Learning (VFL):} 
\begin{equation}
    \mathcal X_i\neq\mathcal X_j, \mathcal Y_i\neq\mathcal Y_j, \mathcal I_i=\mathcal I_j, \forall \mathcal D_i,\mathcal D_j,i \neq j
\end{equation}

\textbf{Federated Transfer Learning (FTL):} 
\begin{equation}
    \mathcal X_i\neq\mathcal X_j, \mathcal Y_i\neq\mathcal Y_j, \mathcal I_i\neq\mathcal I_j, \forall \mathcal D_i,\mathcal D_j,i \neq j
\end{equation}

Here, $\mathcal X$ denotes the feature space, $\mathcal Y$ denotes the label space, and $\mathcal I$ denotes the sample ID space. $\mathcal D_k$ is the dataset owned by the $k$-th party: $\mathcal D_k=(\mathcal X_k, \mathcal Y_k, \mathcal I_k)$.

Besides, the training protocol of HFL also differs from that of VFL. In HFL, each client only keeps its data local but exchanges model parameters or gradients with a server during training, yielding a shared global model among clients. In contrast, each party in VFL keeps both its data and model local and exchanges intermediate results. The outputs after training are separate local models of individual parties.

\section{Methods}

In this section, we present the few-overlap issue in VFL and illustrate our VFLHLP algorithm.

\begin{figure}[h]
    \centering
    \includegraphics[width=0.6\linewidth]{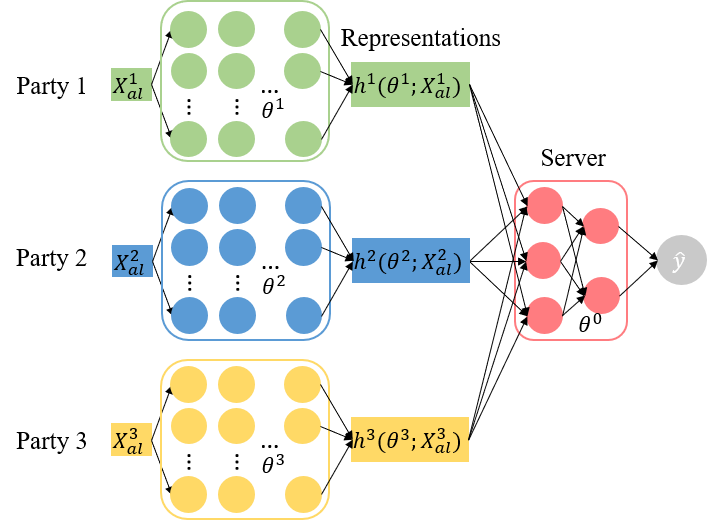}
    \caption{The centralized VFL setting is illustrated by three parties. Party 1 has access to the network and labels on the Server.}
    \label{fig:2}
\end{figure}

\subsection{Problem Definition}

Consider a general VFL setting with $K$ parties. Each party owns different features and only one has labels. To be consistent with other research, the party with labels is called the active party, and the ones holding only features are named the passive parties \cite{liu2022vertical}. Specifically, the $k$-th party has a dataset $X^k=(X^k_{al}, X^k_{nl}), k \in \left\{1,..., K\right\}$, and let only Party 1 own labels $Y=(Y_{al}, Y_{nl})$. 

Regarding conventional VFL, only the small aligned sample set $(Y_{al},$ $X^1_{al},$ $\ldots,$ $X^{K}_{al})$ is used to train a joint model for future collaborative prediction, wasting massive unaligned samples of each party. This severely limits the performance of federated models.

\subsection{Vertical Federated Learning Hybrid Local Pre-training}

In this study, we apply a centralized VFL framework (see Figure \ref{fig:2}) as suggested in \cite{castiglia2022self}. Each party (client) trains its own encoder network $h^k(\cdot)$, parameterized by $\theta^k$, and then sends its representation $r^k=h^k(\theta^k; X^k_{al})$ to the server. The server then concatenates these representations and passes the prediction head $f(\cdot)$, parameterized by $\theta^0$, to make a prediction for classification or regression tasks. Only the active party has access to the predictor $f(\cdot)$ and labels $Y$ on the server.

\begin{figure*}[t]
    \centering
    \includegraphics[width=0.8\linewidth]{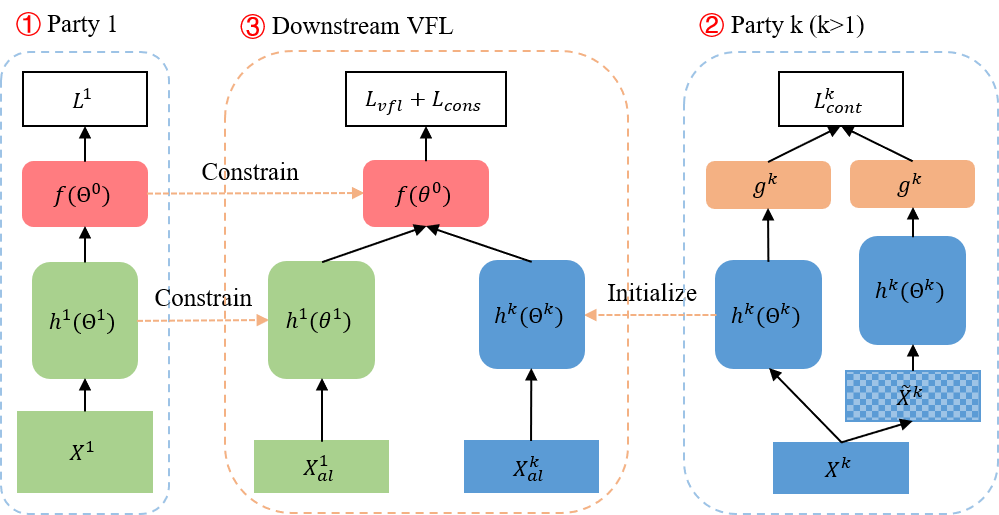}
    \caption{Overview of VFLHLP. VFLHLP contains 3 steps: $\textcircled{1}$ supervised pre-training using local samples of the active party to train local encoder $\Theta^1$ and local prediction head $\Theta^0$; $\textcircled{2}$ self-supervised pre-training using local samples of the passive parties to train individual local encoders $\Theta^k (k>1)$; $\textcircled{3}$ downstream VFL using aligned samples with constraints of  $\Theta^1$ and  $\Theta^0$ and initialization by $\Theta^k (k>1)$.}
    \label{fig:3}
\end{figure*}

We propose a VFLHLP method to address the few-overlap problem. VFLHLP first trains local networks on full local data for individual parties. Specifically, we conduct supervised learning for Party 1 and self-supervised learning for Party $k, k \in \left\{2,..., K\right\}$. Then, we apply these pre-trained networks to improve the federated learning on aligned data.

The core of VFLHLP is to learn effective local networks to enhance downstream federated learning. Therefore, it contains two stages: local pre-training and downstream VFL. 

At the pre-training stage, we implement different strategies for the active party and passive ones. We carry out supervised learning for the active party to exploit local features and valuable labels to obtain well performing encoder and prediction networks (Step $\textcircled{1}$ of Figure \ref{fig:3}) while applying SSL techniques to train effective encoders for feature extraction from local data of the passive parties (Step $\textcircled{2}$ of Figure \ref{fig:3}). At the downstream VFL stage, we exploit these locally pre-trained networks to improve the performance of federated models on the aligned samples (Step $\textcircled{3}$ of Figure \ref{fig:3}).

Specifically, it contains the following three steps:

(1) Supervised pre-training for the active party;

(2) Self-supervised pre-training for the passive parties;

(3) Downstream VFL improved by the pre-trained local networks.

The step (1) and (2) are to pre-train local networks on individual local data, and thus they can be conducted simultaneously. Then step (3) uses these pre-trained networks to improve the downstream VFL model. Therefore, we call our algorithm Vertical Federated Learning Hybrid Local Pre-training (VFLHLP). Notice that the pre-training stage is carried out locally, differing from the additional cross-party configurations in FedCVT \cite{kang2022fedcvt} and FedHSSL \cite{he2022hybrid} methods. Therefore, VFLHLP intrinsically has a much higher efficiency than FedCVT and FedHSSL.

\subsubsection{Supervised Pre-training} 
\label{sup}
To tackle the waste of features and valuable labels of unaligned samples for the active party ($k=1$), we implement supervised learning on its local data. 
Specifically, we train a local model, which has the same network architecture $h^1(\cdot), f(\cdot)$, parameterized by $\Theta^1, \Theta^0$ respectively, on full local data $(X^1, Y)$. With accessible labels, the active party conducts supervised learning to train the model by minimizing the objective:
\begin{equation}
    L^1=F(f(\Theta^0;h^1(\Theta^1;X^1)),Y),
\end{equation}
where, $L^1$ denotes the task loss (e.g., classification task), and $F(\cdot)$ denotes the loss function (e.g., cross entropy loss).

This step generates well-trained weights for the representation and prediction networks of the active party, as they leverage abundant local data with valuable labels. 
They will serve as prior knowledge to be transferred to the sub-model for the active party in the downstream VFL process.

\subsubsection{Self-supervised Pre-training} 
\label{ssl}
To tackle feature waste of unaligned samples for the passive parties ($k>1$), we apply SSL techniques for their local features, as they cannot access labels directly even for aligned samples.
The core of SSL methods is to pre-train a local encoder network by generating pseudo labels from local features in pretext tasks. 
Specifically, we apply Scarf algorithm \cite{bahri2021scarf} as it is simple, versatile, and effective for the scenario of limited labeled data. Scarf applied the idea of contrastive learning \cite{jaiswal2020survey, le2020contrastive}. It generates $\tilde{X}^k$ by corrupting a random subset of features. Then we pass both $X^k$ and $\tilde{X}^k$ through the encoder network $h^k(\cdot)$, parameterized by $\Theta^k$, and the projection head $g^k(\cdot)$ to obtain $z^k$ and $\tilde{z}^k$ respectively. We maximize the similarity between $z^k$ and $\tilde{z}^k$ for the same sample and minimize those of different samples via the InfoNCE contrastive loss, see Eq. (\ref{eq:lct}). By this means, we obtain well performing encoder networks of all passive parties $h^k(\Theta^k)$ for $k, k \in \left\{2, \ldots, K\right\}$.
\begin{gather}
    s^k_{i,j}=\frac{z^k_i \tilde{z}^k_j}{||z^k_i||_2 \cdot ||z^k_j||_2}, for i,j \in N \\
    L_{cont}^k=\frac{1}{N} \sum_{i=1}^N -log(\frac{exp(s^k_{i,i}/\tau)}{\frac{1}{N}\sum_{j=1}^N exp(s^k_{i,j}/\tau)})
    \label{eq:lct}
\end{gather}

\begin{algorithm}[t]
  \caption{Downstream stage of VFLHLP algorithm}
  \label{algo}
  \textbf{Initialize:} $\theta^0_0, \theta^1_0, \theta^k_0=\Theta^k_0 (k>1)$\
  
  \For{$t={0,\ldots,T-1}$}{
    Parties choose randomly sampled mini-batch: $x^k_{al} \in X^k_{al}$\
    
    \For{party $k \in \left\{1,\ldots,K\right\}$ (in parallel)}{
      Party $k$ computes $r^k = h^k(\theta^k_t;x^k_{al})$
      
      Party $k$ sends local representation $r^k$ to server\  
    }
    Server samples corresponding mini-batch: $y^k_{al} \in Y^k_{al}$

    Server computes $L$ according to Eq. (\ref{eq:l})
    
    Server updates $\theta^0_{t+1}=\theta^0_t-\eta_1 \nabla_{\theta^0} L$
    
    Server computes and sends $\nabla_{r^k} L$ to Party $k$
    
    \For{party $k \in \left\{1,\ldots,K\right\}$ (in parallel)}{
      Party $k$ computes $\nabla_{\theta^k} L = \nabla_{\theta^k} r^k \nabla_{r^k} L$
      
      Party $k$ updates $\theta^k_{t+1} =\theta^k_t - \eta_2 \nabla_{\theta^k} L$\
    }
  }
  \textbf{Return:} $\theta^k$, for $k \in \left\{0,\ldots,K\right\}$\
\end{algorithm}

\subsubsection{Downstream Federated Learning}
\begin{figure}[h]
    \centering
    \includegraphics[width=0.8\linewidth]{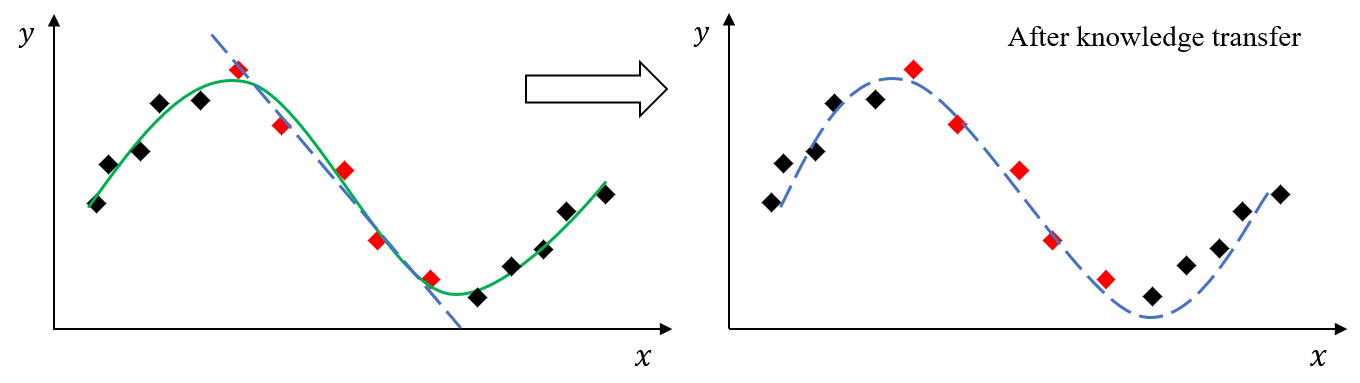}
    \caption{The illustration of knowledge transfer for the active party. Scatters are data samples with red ones as aligned samples and black ones as unaligned samples. The green solid line is the model trained by all local data while the dashed blue line in the left panel represents the model trained by only aligned data. After transferring knowledge from the green solid line to the blue dashed line, we can obtain a more accurate model shown by a new dashed blue line in the right panel.}
    \label{fig:4}
\end{figure}

Finally, we exploit priorly obtained local information to improve federated models at the downstream supervised VFL stage.
Specifically, each party trains a local encoder network on the aligned sample set with initialized network weights as $\Theta^k$ from Step $\textcircled{2}$ of Figure \ref{fig:3} except the active party, and then sends its representation $r^k=h^k(\theta^k; X^k_{al})$ to the server. The server combines these representations $H=\left\{r^1,\ldots,r^K\right\}$ and trains the prediction network by minimizing the objective: $L_{vfl}=F(f(\theta^0;H),Y_{al})$. The initial weights $\Theta^k$ derived from Step \textcircled{2} of Figure \ref{fig:3} provide well starting for downstream fine-tuning on the small aligned sample set for the passive parties. 
\begin{gather}
    L = L_{vfl} + \beta L_{cons} \label{eq:l} \\
    L_{cons} = 0.5 ((\theta^1 - \Theta^1)^2 + (\theta^0_{s} - \Theta^0)^2) \label{eq:lcs}
\end{gather}

Regarding the active party, we take the idea of knowledge transfer, exploiting the prior knowledge derived from Step \textcircled{1} of Figure \ref{fig:3} to constrain the sub-model during federated learning. A naive assumption for this is that a model trained on a huge dataset has much less bias than the one trained on its small sampled subset, as illustrated in Figure \ref{fig:4}. We use the pre-trained model weights $\Theta^1$ and $\Theta^0$ to online adjust $\theta^1$ and $\theta^0_{s}$ by introducing constrain loss $L_{cons}$ into the objective during federated training, see Eq. (\ref{eq:lcs}). Here, $\theta^0_{s}$ represents the sub-model of the prediction for the active party, and $\beta$ denotes the coefficient for the constraint loss. Algorithm \ref{algo} gives a full view of the downstream VFL stage in VFLHLP.


\section{Experiments and Results}
In this section, we conduct experiments on two real-world advertising datasets called Avazu \cite{avazu} and Criteo \cite{criteo} to demonstrate our method. We compare our method VFLHLP with baselines trained in other scenarios displayed below:
\begin{itemize}
\item{\textbf{Local-A.}} The model is trained on the local data of the active party.
\item{\textbf{VFL.}} The vanilla VFL is trained on the aligned samples.
\end{itemize}

Furthermore, we also compared our method with other baseline algorithms and implemented an ablation study on our method.
To demonstrate the robustness of our method over the amount of aligned data, we also vary the number of aligned samples while fixing the size of local samples. 
For all the experiments we use Area-Under-Curve (AUC) as the metric to evaluate model performance and repeat each of them with five different seeds to obtain the average results.

\begin{table}
  \caption{Dataset statistics and model architecture. "\#fields" denotes the number of data fields in each party. "\#dim" denotes the dimension of the input features after embedding. "|" segments values for different parties.}
  \label{tab:1}
  \centering
  \begin{tabular}{l|c|c}
    \toprule
    \textbf{Item} & \textbf{Avazu} & \textbf{Criteo}\\
    \midrule
    \#total & 100000 & 80000\\
    \#parties & 3 & 2 \\
    \#aligned & 250 $\sim$  8000 & 200 $\sim$ 2000\\
    \#test & 20000 & 10000 \\
    \midrule
    \midrule
    \#fields & 7 | 7 | 8 & 13 | 26 \\
    \#dim & 56 | 56 | 64 & 13 | 416 \\
    \midrule
    \midrule
    encoder & 64 $\rightarrow$ 64 $\rightarrow$ 16 & 256 $\rightarrow$ 128 $\rightarrow$ 64 $\rightarrow$ 16 \\
    prediction head & 48 $\rightarrow$ 1 & 32 $\rightarrow$ 1 \\
    \bottomrule
    \end{tabular}
\end{table}

\subsection{Experiments on Avazu}
\label{ava}
Avazu is a public dataset for click-through rate (CTR) prediction. It only contains categorical features with 23 fields in total. We removed the ``id'' field and converted the ``hour'' field into the 0-23 range before inputting them into the model. We transform the categorical features into embeddings with a fixed dimension of 8 in this experiment. 
For computational efficiency, we randomly select 100000 samples for training and 20000 samples for testing from the original dataset. 

In this experiment, we set a three-party scenario with each party owning 50000 training samples and 20000 aligned test samples. The number of aligned training samples ranges from 250 to 8000, accounting for 0.5\% to 16\% of local samples. A comprehensive overview of the vertically partitioned datasets and the associated model architecture can be found in Table \ref{tab:1}.

As shown in Table \ref{tab:2}, VFLHLP outperforms vanilla VFL by a large margin. 
For instance, FedHSSL enhances model AUC by 8.9\% with 250 aligned samples and 3.5\% with 8000 samples compared with vanilla VFL. The AUC improvement declines with an increased amount of aligned samples, as the unique information from unaligned samples gets less. Notably, VFLHLP achieves better performance with much fewer aligned samples compared to vanilla VFL. We observe that with only 250 aligned samples VFLHLP reaches an AUC of 0.694 surpassing the performance of vanilla VFL using 8000 aligned samples. Vanilla VFL still underperforms Local-A with even 8000 aligned samples. In contrast, VFLHLP catches Local-A with 4000 aligned samples and outperforms it with 8000 ones.


\begin{table*}[t]
  \caption{Test AUC ($\uparrow$) comparison of VFLHLP to vanilla VFL on Avazu dataset with a varying aligned sample size. $\Delta$ is the performance boost of VFLHLP compared to vanilla VFL. Local-A is trained on 50000 local samples of the active party.}
  \label{tab:2}
  \centering
  \scalebox{0.95}{
  \begin{tabular}{lcccccc}
    \toprule
    \multirow{2}{*}{Model} & \multicolumn{6}{c}{aligned number}  \\
    \cmidrule(r){2-7}
    \multirow{2}{*}{} & 250 & 500 & 1000 & 2000 & 4000 & 8000 \\
    \midrule
    \midrule
    Local-A & 0.719 $\pm$ 0.002 & 0.719 $\pm$ 0.002 & 0.719 $\pm$ 0.002 & 0.719 $\pm$ 0.002 & 0.719 $\pm$ 0.002 & 0.719 $\pm$ 0.002 \\
    VFL & 0.605 $\pm$ 0.013 & 0.622 $\pm$ 0.016 & 0.636 $\pm$ 0.021 & 0.659 $\pm$ 0.014 & 0.670 $\pm$ 0.006 & 0.688 $\pm$ 0.015 \\
    VFLHLP & \textbf{0.694} $\pm$ 0.012 & \textbf{0.703} $\pm$ 0.008 & \textbf{0.710} $\pm$ 0.004 & \textbf{0.717} $\pm$ 0.004 & \textbf{0.719} $\pm$ 0.002 & \textbf{0.723} $\pm$ 0.002 \\
    $\Delta$ VFLHLP & $\uparrow$ 0.089 & $\uparrow$ 0.081 & $\uparrow$ 0.074 & $\uparrow$ 0.058 & $\uparrow$ 0.049 & $\uparrow$ 0.035\\
    \bottomrule
  \end{tabular}
  }
\end{table*}

\subsection{Experiments on Criteo}
Criteo is another public well-known CTR dataset. It is composed of 26 categorical features and 13 numerical features. We transformed categorical features into embeddings with a fixed dimension of 16 and scaled the numerical features into the 0-1 range during the pre-processing stage. 

To emulate the VFL setting, we establish a two-party scenario where the active party owns categorical features and labels while the passive party holds numerical ones. We randomly select 100000 samples from the original dataset to reduce the computational complexity. Each party has 40000 training samples, 10000 aligned validation samples, and 10000 aligned test samples. The detailed configuration of this experiment is summarized in Table \ref{tab:1}.

Table \ref{tab:3} displays the AUC comparison between VFLHLP and baselines with a varying number of aligned training samples from 200 to 2000, accounting for 0.5\% to 5\% of local samples. We observe that VFLHLP outperforms Local-A significantly and surpasses vanilla VFL remarkably. 
With 200 aligned samples, it raises AUC by 1.1\% over Local-A and by 11.5\% over vanilla VFL. When the aligned sample size increases to 2000, the improvement in AUC by VFLHLP is 2.7\% over Local-A and 8.1\% over vanilla VFL.
Similar to the results for Avazu, the performance gain over vanilla VFL declines with a larger aligned sample size. Moreover, the improvement over Local-A presents an increasing trend since the feature information from passive parties grows.
Furthermore, it is notable that VFLHLP outperforms Local-A with only 200 aligned samples, which highlights the significant impact of the numerical features owned by the passive party.

\begin{table*}[t]
  \caption{Test AUC ($\uparrow$) comparison of VFLHLP to vanilla VFL on Criteo dataset with a varying aligned sample size. $\Delta$ is the performance boost of VFLHLP compared to vanilla VFL. Local-A is trained on 40000 local samples of the active party.}
  \label{tab:3}
  \centering
  \scalebox{0.95}{
  \begin{tabular}{lcccccc}
    \toprule
    \multirow{2}{*}{Model} & \multicolumn{6}{c}{aligned number}  \\
    \cmidrule(r){2-7}
    \multirow{2}{*}{} & 200 & 400 & 600 & 800 & 1000 & 2000\\
    \midrule
    \midrule
    Local-A & 0.668 $\pm$ 0.006 & 0.668 $\pm$ 0.006 & 0.668 $\pm$ 0.006 & 0.668 $\pm$ 0.006 & 0.668 $\pm$ 0.006 & 0.668 $\pm$ 0.006\\
    VFL & 0.564 $\pm$ 0.012 & 0.575 $\pm$ 0.009 & 0.585 $\pm$ 0.012 & 0.588 $\pm$ 0.010 & 0.594 $\pm$ 0.008 & 0.614 $\pm$ 0.004\\
    VFLHLP & \textbf{0.679} $\pm$ 0.009 & \textbf{0.682} $\pm$ 0.007 & \textbf{0.684} $\pm$ 0.009 & \textbf{0.691} $\pm$ 0.006 & \textbf{0.690} $\pm$ 0.005 & \textbf{0.695} $\pm$ 0.006\\
    $\Delta$ VFLHLP & $\uparrow$ 0.115 & $\uparrow$ 0.107 & $\uparrow$ 0.099 & $\uparrow$ 0.103 & $\uparrow$ 0.096 & $\uparrow$ 0.081 \\
    \bottomrule
  \end{tabular}
  }
\end{table*}

\subsection{VFLHLP with Baselines}
\label{sec:bs}

We further compared our approach with several well-known baseline algorithms: 
\begin{itemize}
\item{\textbf{FedSplitNN.}} This is vanilla VFL implemented in FedHSSL's framework. We name it originally as in \cite{he2022hybrid} for consistency.
\item{\textbf{FedLocalSSL.}} This series applies various SSL methods for both the active and passive parties to enhance the effectiveness of encoder networks for downstream fine-tuning on small aligned data. It contains three representative SSL methods: Simple Siamese (SimSiam) \cite{chen2021exploring}, Bootstrap Your Own Latent (BYOL) \cite{grill2020bootstrap}, and Momentum Contrast (MoCo) \cite{he2020momentum}. Correspondingly, we name them FedLocalSimSiam, FedLocalBYOL, and FedLocalMoCo, respectively. These algorithms cover the idea of SS-VFL \cite{castiglia2022self} and VFLFS \cite{feng2022vertical}.
\item{\textbf{FedHSSL.}} This is supposed to be a state-of-the-art algorithm series \cite{he2022hybrid} which surpasses the performance of many well-known methods (e.g., FedCVT \cite{kang2022fedcvt}). 
\end{itemize}

We compared VFLHLP with these baselines on the same Avazu sub-dataset as described in Section \ref{ava}. Unlike the former three-party setting, we partitioned the dataset into only two parts to keep consistent with \cite{he2022hybrid}. Each party also owns 50000 training samples and 20000 aligned test samples. Notice that the active party has full labels of local samples. We integrated our algorithm into the framework of FedHSSL and applied the same model architectures and hyperparameters as the ones in \cite{he2022hybrid}. The details of this experimental configuration are referenced in \cite{he2022hybrid}. 

\begin{table*}
  \caption{Test AUC ($\uparrow$) comparison of VFLHLP to baselines on Avazu dataset with a varying aligned sample size.}
  \label{tab:4}
  \centering
  \begin{tabular}{lccccc}
    \toprule
    \multirow{2}{*}{Model} & \multicolumn{5}{c}{aligned number}  \\
    \cmidrule(r){2-6}
    \multirow{2}{*}{} & 200 & 400 & 600 & 800 & 1000 \\
    \midrule
    \midrule
    FedSplitNN & 0.592 $\pm$ 0.013 & 0.607 $\pm$ 0.014 & 0.611 $\pm$ 0.005 & 0.624 $\pm$ 0.009 & 0.621 $\pm$ 0.012 \\
    FedLocalSimSiam & 0.607 $\pm$ 0.013 & 0.625 $\pm$ 0.004 & 0.634 $\pm$ 0.007 & 0.634 $\pm$ 0.008 & 0.633 $\pm$ 0.014 \\
    FedLocalBYOL & 0.606 $\pm$ 0.008 & 0.621 $\pm$ 0.010 & 0.630 $\pm$ 0.009 & 0.634 $\pm$ 0.019 & 0.632 $\pm$ 0.012 \\
    FedLocalMoCo & 0.608 $\pm$ 0.014 & 0.620 $\pm$ 0.009 & 0.630 $\pm$ 0.015 & 0.625 $\pm$ 0.016 & 0.633 $\pm$ 0.007 \\
    FedHSSL-SimSiam & 0.627 $\pm$ 0.012 & 0.631 $\pm$ 0.011 & 0.637 $\pm$ 0.016 & 0.639 $\pm$ 0.013 & 0.639 $\pm$ 0.015 \\
    FedHSSL-BYOL & 0.617 $\pm$ 0.015 & 0.633 $\pm$ 0.004 & 0.634 $\pm$ 0.013 & 0.639 $\pm$ 0.008 & 0.642 $\pm$ 0.008 \\
    FedHSSL-MoCo & 0.622 $\pm$ 0.017 & \textbf{0.637} $\pm$ 0.010 & 0.638 $\pm$ 0.012 & 0.649 $\pm$ 0.007 & 0.649 $\pm$ 0.012 \\
    VFLHLP (ours) & \textbf{0.635} $\pm$ 0.009 & 0.629 $\pm$ 0.007 & \textbf{0.656} $\pm$ 0.006 & \textbf{0.661} $\pm$ 0.006 & \textbf{0.662} $\pm$ 0.006 \\
    \bottomrule
  \end{tabular}
\end{table*}

As illustrated in Table \ref{tab:4}, our approach VFLHLP generally generates superior performance, outperforming FedSplitNN, FedLocalSSL-series, and FedHSSL-series methods. For example, with 200 aligned samples VFLHLP rises up AUC by 2.7\% and 0.8\% compared to FedLocalMoCo (the best in the FedLocalSSL series) and FedHSSL-SimSiam (the best in the FedHSSL series), respectively. 
We notice that the superiority of VFLHLP compared to FedHSSL is more significant with the aligned sample size larger than 400. In particular, VFLHLP surpasses FedHSSL-MoCo (the best in the FedHSSL series) by 1.8\% in AUC with 600 aligned samples, accounting for 1.2\% of local data volume.

\subsection{Ablation Study}
\begin{figure}[h]
    \centering
    \includegraphics[width=0.65\linewidth]{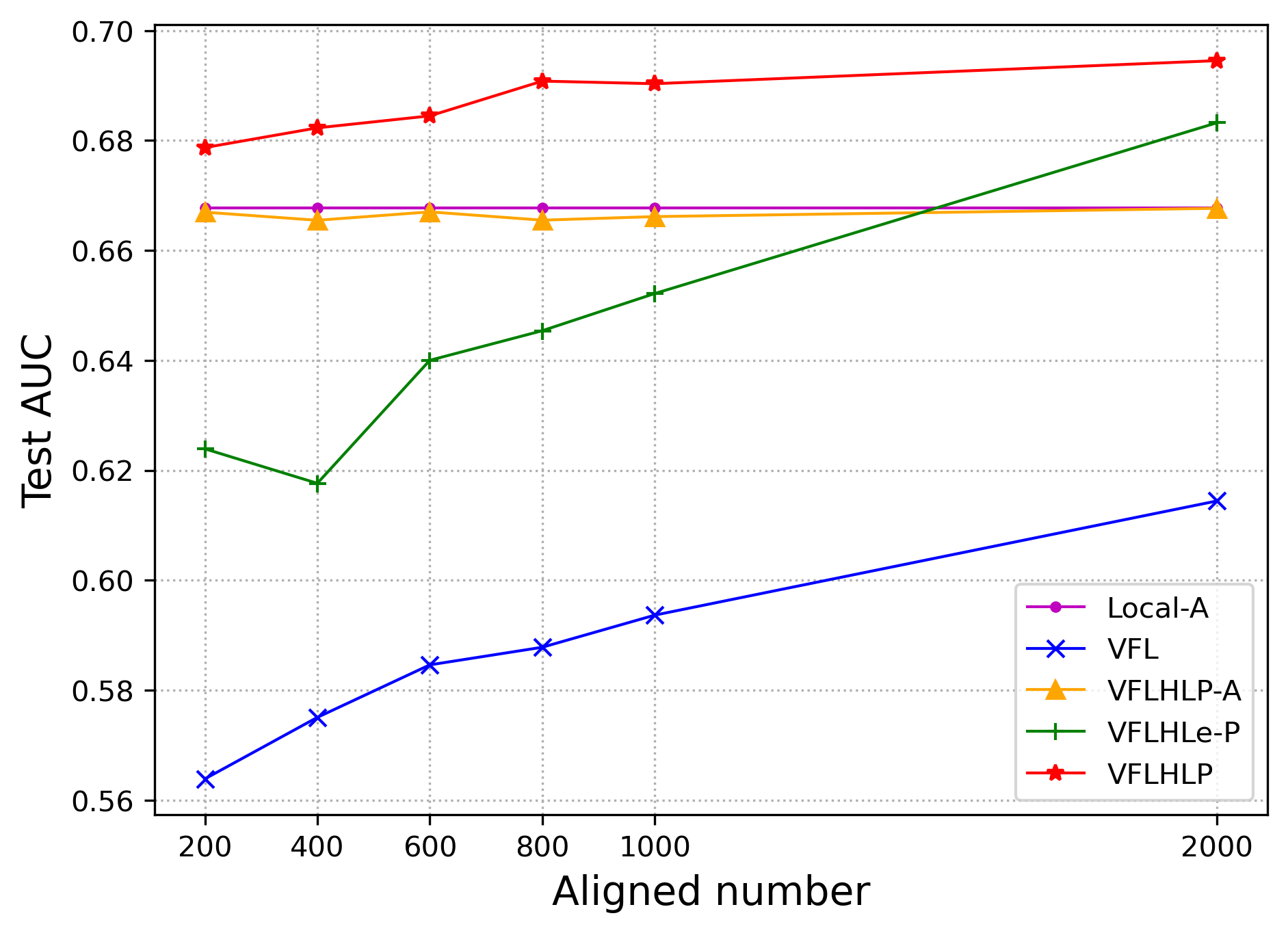}
    \caption{Test AUC ($\uparrow$) comparison of VFLHLP to baselines on Criteo dataset with a varying aligned sample size.}
    \label{fig:5}
\end{figure}

VFLHLP exploits the pre-trained networks by local samples to enhance downstream federated model performance, including the pre-trained model of the active party and the pre-trained encoders of the passive parties. In this section, we study how each side contributes to the performance of VFLHLP. Specifically, we measure the following scenarios:
\begin{itemize}
    \item{\textbf{VFLHLP-A.}} Use only the pre-trained model of the active party to constrain the downstream VFL. In other words, only use the local data of the active party and apply the idea of knowledge transfer.
    \item{\textbf{VFLHLP-P.}} Use only the pre-trained encoders of passive parties for the initialization of the downstream sub-models for the passive parties. In other words, only use the local data of the passive parties and apply SSL methods.
    \item{\textbf{VFLHLP.}} Use all available local data and apply both techniques.
\end{itemize}

We conducted the ablation study on Criteo with an aligned sample size varying from 200 to 2000, accounting for 0.5\% to 5\% of the local sample size. The experimental results are shown in Figure \ref{fig:5}. We observe that all scenarios present a performance boost compared with vanilla VFL. Among them, VFLHLP has the most prominent performance improvement by more than 10\%. The contribution of the active party reflected by the difference between VFLHLP and vanilla VFL is significantly higher than that of the passive party with an aligned sample size of less than 1000. 
Furthermore, we notice that the contribution of the active party declines with the increased number of aligned samples while the one for the passive party is almost consistent. This suggests that while the active party's contribution is significant, the passive party's role becomes increasingly influential with larger volumes of aligned samples. In particular, the contribution by the passive party significantly surpasses the one by the active party with 2000 aligned samples.

\section{Conclusions}
We propose VFLHLP, a novel approach to address the few-overlap problem in the sample space widely existing in real-world VFL applications. VFLHLP utilizes local samples (both aligned and unaligned) of participating parties to boost the performance of federated models. It utilizes the idea of knowledge transfer, exploiting the pre-trained local model of the active party to constrain downstream VFL, and applies the Scarf technique, using the pre-trained encoders of passive parties to initialize the sub-models of passive parties during the VFL stage. The experimental results show that VFLHLP outperforms baselines significantly. Besides, the ablation study demonstrates the significance of each technique implemented in VFLHLP.


\bibliographystyle{unsrtnat}
\bibliography{vfl_paper}



\end{document}